\documentclass[conference,letterpaper]{IEEEtran}

\usepackage[T1]{fontenc}
\usepackage{cite}
\usepackage{amsmath,amssymb,amsfonts}
\usepackage{algorithmic}
\usepackage{graphicx}
\usepackage{textcomp}
\usepackage{xcolor}
\usepackage{comment}
\usepackage{hhline}
\usepackage{multirow}

\def\BibTeX{{\rm B\kern-.05em{\sc i\kern-.025em b}\kern-.08em
    T\kern-.1667em\lower.7ex\hbox{E}\kern-.125emX}}
\begin{document}

\title{MS-RFD: Multi-Signal Release Frame Detection in Hammer Throw from Reconstructed 3D Trajectories}

\author{
\IEEEauthorblockN{
Ahmed Endris Hasen\textsuperscript{1,*},
Nikolaos Passalis\textsuperscript{2},
Tomi Vänttinen\textsuperscript{3},
Jenni Raitoharju\textsuperscript{1}
}
\IEEEauthorblockA{
\textsuperscript{1}Faculty of Information Technology, University of Jyväskylä, Jyväskylä, Finland\\
\textsuperscript{2} Faculty of Sciences, Aristotle University of Thessaloniki, Thessaloniki, Greece\\
\textsuperscript{3}Finnish Institute of High Performance Sport KIHU, Jyväskylä, Finland\\
\textsuperscript{*}Corresponding author: ahmed.e.hasen@jyu.fi\\
Email: ahmed.e.hasen@jyu.fi, passalis@csd.auth.gr, tomi.vanttinen@kihu.fi, jenni.k.raitoharju@jyu.fi
}
}

\maketitle

\begin{abstract}
Recent advances in artificial intelligence and computer vision are reshaping sports performance analysis by enabling automated detection, tracking, and performance analysis. In hammer throw, performance is strongly determined by the kinematic conditions at release, particularly release speed, release angle, and release height. However, identifying the release instant from video typically requires manual frame-by-frame inspection, which is subjective and cumbersome in real-world training scenarios. In this paper, we present a fully automatic multi-signal release frame detection (MS-RFD) method for hammer throw using reconstructed 3D hammer trajectories. The proposed method integrates four complementary kinematic signals: speed dynamics, angular velocity transition, radial distance relative to the rotation center, and post-release trajectory linearity. These signals are fused to score and verify candidate release frames. MS-RFD is evaluated through the throwing-distance estimation error obtained from the release parameters estimated at the detected frame. An ablation study analyzes the contribution of each signal and compares alternative candidate selection strategies. The results show that speed dynamics and radial expansion provide the strongest signals for release frame detection, while angular velocity and post-release linearity provide smaller refinements. 
\end{abstract}


\section{Introduction}

Recent advances in computer vision and artificial intelligence are rapidly transforming sports performance analysis~\cite{naik2022comprehensive}, enabling tasks such as player and ball detection~\cite{buric2019adapting}, tracking and trajectory based analysis~\cite{torres2022tracking}, and action or event detection~\cite{xu2025deep}. In biomechanics-driven sports, such as hammer throw, performance analysis often depends on accurate estimation of release parameters, including release speed, release angle, and release height~\cite{castaldi2022biomechanics},~\cite{ding2025kinematic}. These parameters define the subsequent ballistic motion and strongly influence the final throwing distance~\cite{dapena2003prediction}. Previous biomechanical studies have examined release conditions and the radius of curvature during turns as important quantities in hammer throw analysis~\cite{dapena2003prediction},~\cite{murofushi2005radius}.

Video-based analysis provides a practical way to estimate release parameters and predict throwing distance, particularly in training environments where direct measurement may be difficult. In our previous work~\cite{hasen2026hammer}, we introduced a video-based hammer throw distance estimation and trajectory analysis pipeline that combines synchronized multi-view video, DLT-based camera calibration, hammer-head detection, and stereo triangulation to reconstruct 3D trajectories from video inputs. Despite the promising results, automatic release frame identification remained a limiting component of the pipeline. However, identifying the release frame is a critical step for estimating release parameters, extracting meaningful biomechanical insights and enabling trajectory-based distance estimation from videos.

In our previous work~\cite{hasen2026hammer}, release frame identification was mainly performed manually through frame-by-frame video inspection, which is cumbersome and dependent on expert judgment. This limits scalability and introduces subjectivity into performance analysis workflows. A simple automatic baseline was also included, where the release frame was estimated from a weighted combination of speed and angular velocity computed from the reconstructed 3D trajectory. Although this provided an initial automatic solution, the accuracy was limited. The method relied only on two local kinematic cues and did not explicitly model the broader transition from rotational motion to ballistic flight. These limitations motivate the need for a more systematic and reliable automatic release frame detection method based on the reconstructed 3D trajectories.

In this paper, we focus on release frame detection and propose a multi-signal approach operating on reconstructed 3D hammer trajectories. The proposed multi-signal release frame detection (MS-RFD) method integrates multiple physics-informed temporal signals, including speed dynamics, radial distance variation relative to the rotational center, angular velocity transition, and post-release trajectory linearity. These signals are combined within a unified scoring framework and followed by biomechanical candidate validation and final-frame selection. Unlike the previous method that relies on limited signal rules, the proposed method uses multiple complementary cues to better detect the transition from rotational motion to ballistic flight. The main contributions of this paper are as follows:

\begin{itemize}
    \item We formulate hammer throw release frame identification from multi-view videos as an automatic release frame detection problem based on the reconstructed 3D trajectories.
    
    \item We propose MS-RFD, a multi-signal release frame detection method that combines speed dynamics, radial expansion, angular velocity transition, and post-release trajectory linearity.

    \item We conduct an ablation study to analyze the contribution of individual signals and to compare alternative candidate selection strategies for release frame detection.

    \item We evaluate the proposed method using the biomechanical consistency of the release parameters and throw distance estimation accuracy.
\end{itemize}

\section{Method}
\label{sec:method}

\subsection{Overview of the Proposed Method}

The proposed method builds on the 3D trajectory reconstruction pipeline introduced in~\cite{hasen2026hammer}. Given synchronized dual-view video, the hammer head is detected in each camera view, cameras are calibrated using Direct Linear Transformation (DLT), and the 3D hammer positions are reconstructed through stereo triangulation and the resulting trajectory is smoothed before release analysis to reduce noise and improve temporal consistency.

Let
\[
\mathcal{T}=\{\mathbf{p}_1,\mathbf{p}_2,\ldots,\mathbf{p}_n\}
\]
denote the reconstructed 3D hammer trajectory, where $\mathbf{p}_i=(x_i,y_i,z_i)\in\mathbb{R}^3$ represents the estimated hammer position at frame $i$. The objective is to estimate the release frame \(r^*\), corresponding to the transition from rotational motion to ballistic flight. This release-frame detection problem can be written generally as
\[
r^* = F(\mathcal{T}),
\]
where \(F\) is an automatic release-detection function that maps the reconstructed trajectory to a release-frame index. 

\begin{figure*}[htb]
  \centering
  \includegraphics[width=17.0cm]{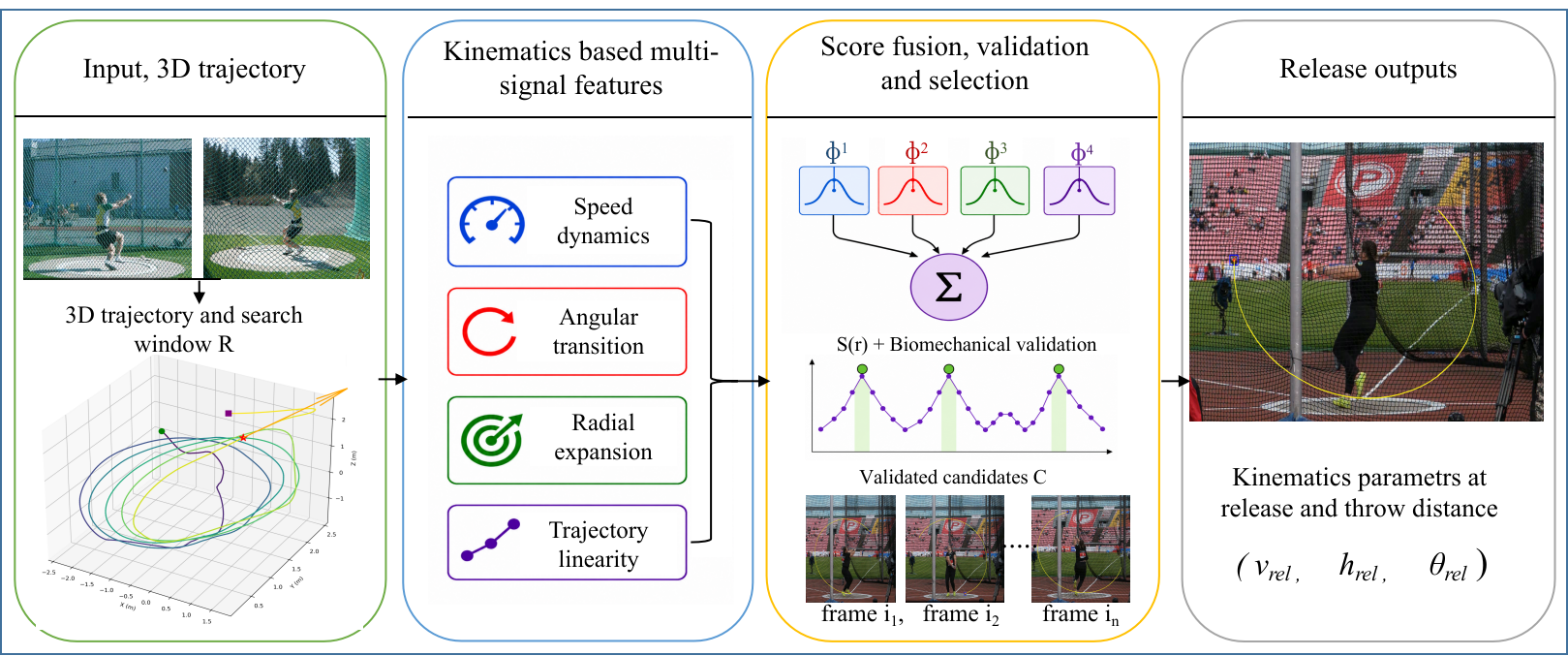}
  \caption{Overview of the proposed MS-RFD framework for automatic release-frame detection from reconstructed 3D hammer trajectories.}
  \label{fig:method1}
\end{figure*}

Fig.~\ref{fig:method1} illustrates the proposed MS-RFD framework. In this work, \(F\) is implemented through three main stages: candidate search-window selection, multi-signal release scoring, and candidate validation/selection. The method first selects a candidate search window \(\mathcal{R}\) from the final rotational phase of the trajectory. Since release is expected to occur during the last complete rotational cycle, this window is estimated from the radial-distance behavior of the reconstructed trajectory and placed around the final circular-motion region. Candidate frames inside \(\mathcal{R}\) are then evaluated using four normalized kinematic signals: speed dynamics, angular transition, radial expansion, and post-release linearity. These signals are fused into a multi-signal release score \(S(r)\), which is used to rank candidate frames within the search window as described in Section~\ref{ssec:scroing}. Candidate frames are then validated using biomechanical release-parameter constraints, and the final release frame is selected from the validated candidate set using one of the selection strategies described in Section~\ref{ssec:validation}.

\subsection{Multi-Signal Release Scoring}
\label{ssec:scroing}
The release event is characterized by a transition from circular motion to ballistic motion. To capture this, four complementary kinematic cues are extracted from the reconstructed trajectory: speed dynamics, angular transition, radial expansion, and post-release linearity. The method evaluates multiple candidate frames \(r\) within the search window \(\mathcal{R}\).



For each signal, unnormalized score \(g_j(r)\) is first computed from the trajectory for each candidate frame \(r\). The final signal score \(\phi_j(r)\) is then obtained by normalizing the signal values within the search window:
\[
\phi_j(r)=\mathcal{N}_{\mathcal{R}}(g_j(r)),
\]
where \(\mathcal{N}_{\mathcal{R}}(\cdot)\) rescales the values in \(\mathcal{R}\) to \([0,1]\). Thus, larger \(\phi_j(r)\) values indicate stronger release evidence.

\subsubsection{Speed Dynamics ($\boldsymbol{\phi}_1$)}

Velocity is computed using finite differences. A central-difference is used to obtain a smoother estimate:
\begin{equation}
\mathbf{v}_i=\frac{\mathbf{p}_{i+1}-\mathbf{p}_{i-1}}{2\Delta t}, 
\quad \Delta t=\frac{1}{f_{\text{fps}}}.
\end{equation}
The speed magnitude is $s_i=\|\mathbf{v}_i\|$. 
Hammer speed typically increases through the turns and reaches a high value near release~\cite{murofushi2007hammer},~\cite{bandou2006relationship},~\cite{brice2011analysis}. The speed-dynamics signal therefore captures positive increases in smoothed hammer speed near release while reducing sensitivity to short-term noise and fluctuations in the reconstructed trajectory. The speed signal is defined from the positive temporal increase of the smoothed speed:
\[
g_1(r)=\max\left(0,\frac{d\tilde{s}_r}{dt}\right),
\qquad
\phi_1(r)=\mathcal{N}_{\mathcal{R}}(g_1(r)),
\]
where \(\tilde{s}_r\) is the smoothed speed at candidate frame \(r\).

\subsubsection{Angular Transition ($\boldsymbol{\phi}_2$)}

The transition from rotational to more linear motion can also be characterized through changes in the direction of velocity. The angular change between consecutive velocity vectors is
\begin{equation}
\theta_i = \arccos\left(\frac{\mathbf{v}_{i-1} \cdot \mathbf{v}_i}{\|\mathbf{v}_{i-1}\| \|\mathbf{v}_i\|}\right)
\end{equation}

During rotation, the velocity direction changes rapidly. After release, the horizontal trajectory becomes more linear, and directional change decreases. This signal evaluates the transition comparing angular behavior before and after each candidate frame:
\[
g_2(r)=\max\left(0,\bar{\theta}^{\,\mathrm{pre}}_r-\bar{\theta}^{\,\mathrm{post}}_r\right),
\qquad
\phi_2(r)=\mathcal{N}_{\mathcal{R}}(g_2(r)),
\]
where \(\bar{\theta}^{\,\mathrm{pre}}_r\) and \(\bar{\theta}^{\,\mathrm{post}}_r\) denote the mean angular change before and after \(r\), respectively.

\subsubsection{Radial Distance Expansion ($\boldsymbol{\phi}_3$)}

During the turns, the hammer follows an approximately circular path in the horizontal plane. The rotation center $\mathbf{c}$ is estimated from stable mid-trajectory points in the horizontal $xy$ plane using a robust RANSAC~\cite{fischler1981random} based approximation. The radial distance is defined as
\begin{equation}
\rho_i=\|\mathbf{p}_i^{xy}-\mathbf{c}\|,
\end{equation}
where $\mathbf{p}_i^{xy}=(x_i,y_i)$ denotes the horizontal projection. At release, the hammer departs from the circular path, resulting in sustained radial expansion. The radial-expansion signal measures positive consistent outward radial progression over a future window:
\[
g_3(r)=\max\left(0,\rho_{r+L}-\rho_r\right),
\qquad
\phi_3(r)=\mathcal{N}_{\mathcal{R}}(g_3(r)),
\]
where \(L\) is the future window length.

\subsubsection{Post-Release Linearity ($\boldsymbol{\phi}_4$)}

After release, the hammer follows projectile motion, and its trajectory becomes locally linear over a short interval. For each candidate frame \(r\), a line is fitted to a short future segment of the 3D trajectory from \(r\) to \(r+L\), and linearity is evaluated using a coefficient of determination:
\begin{equation}
R^2_r=1-
\frac{\sum_{k=r}^{r+L} \|\mathbf{q}_k-\hat{\mathbf{q}}_k\|^2}
{\sum_{k=r}^{r+L} \|\mathbf{q}_k-\bar{\mathbf{q}}\|^2},
\end{equation}
where \(k\) indexes the frames in the future segment, \(\mathbf{q}_k\) denotes the 3D trajectory point at frame \(k\), \(\hat{\mathbf{q}}_k\) is its projection onto the fitted 3D line, and \(\bar{\mathbf{q}}\) is the segment mean. Higher values of $R^2_r$ indicate stronger linearity and therefore stronger consistency with post-release ballistic motion.
The linearity signal is defined as
\[
g_4(r)=R_r^2,
\qquad
\phi_4(r)=\mathcal{N}_{\mathcal{R}}(g_4(r)).
\]

\subsubsection{Score Fusion}

The multi-signal release score for each candidate frame is computed by combining the normalized signals:
\begin{equation}
S(r)=\sum_{j=1}^{4} w_j \phi_j(r),
\quad \sum_{j=1}^{4}w_j=1,
\end{equation}
where \(S(r)\) is the fused release score for candidate frame \(r\) and $w_j$ is the weight for signal score $\phi_j(r)$. The fused score \(S(r)\) is used to rank candidate frames within the search window \(\mathcal{R}\), with higher values indicating stronger release likelihood. The weights are specified as $w_1=0.40$, $w_2=0.20$, $w_3=0.30$, and $w_4=0.10$, corresponding to speed dynamics, angular transition, radial expansion, and post-release linearity, respectively. The weights were empirically selected based on the relevance of each cue and analyzed through ablation experiments. 

\subsection{Candidate Selection and Validation} 
\label{ssec:validation}
After using the multi-signal score \(S(r)\) to rank candidate frames within the search window \(\mathcal{R}\), the frames with the highest release scores are retained as candidate release frames and are filtered using biomechanical constraints.
To ensure biomechanical consistency and to reduce false detections, candidate frames are retained only if their estimated release speed, release angle, and release height fall within realistic hammer throw ranges:
\begin{equation}
\begin{aligned}
(s_{\min} \leq s_r \leq s_{\max})
&\land 
(\alpha_{\min} \leq \alpha_r \leq \alpha_{\max}) \\
&\land 
(h_{\min} \leq h_r \leq h_{\max}),
\end{aligned}
\end{equation}
where $s_r$, $\alpha_r$, and $h_r$ denote release speed, angle, and height at candidate frame \(r\), respectively. The parameter bounds are chosen according to plausible hammer throw release conditions reported in biomechanics literature~\cite{pavlovic2020biomechanical},~\cite{isele2010biomechanical},~\cite{castaldi2022biomechanics}. If no valid candidate remains, the search window is expanded and the validation step is repeated. 
Let \(\mathcal{C}\subseteq\mathcal{R}\) denote the validated candidate-frame set. Thus, frames in \(\mathcal{C}\) are selected based on the multi-signal score \(S(r)\) and satisfy the biomechanical plausibility constraints. The final release estimate is then obtained from \(\mathcal{C}\) using one of the selection strategies described below.

\subsubsection{Weighted Mean Aggregation (S1)}
Each candidate frame \(r \in \mathcal{C}\) is assigned a weight $\lambda_r$ based on its speed proximity to the robust candidate speed center $\mu_s$:
\begin{equation}
\lambda_r=\exp\left(
-\frac{(s_r-\mu_s)^2}{2\sigma_s^2}
\right).
\end{equation}
where \(s_r\) denotes the speed at candidate frame \(r\), while \(\mu_s\) and \(\sigma_s\) are computed from the speeds of the validated candidates in \(\mathcal{C}\). The weighted mean frame is first computed, and then final release frame is selected as the validated candidate closest to this weighted mean:
\[
r^*=\arg\min_{r\in\mathcal{C}}
\left|
r-
\frac{\sum_{u\in\mathcal{C}}\lambda_u u}
{\sum_{u\in\mathcal{C}}\lambda_u}
\right|,
\]

\subsubsection{Representative Frame Selection (S2)} 
The representative-frame strategy selects the candidate whose speed is closest to the median candidate speed:
\begin{equation}
r^* = \underset{r \in \mathcal{C}}{\arg\min} \, |s_r - \text{median}(\{s_i\}_{i \in \mathcal{C}})|
\end{equation}
This approach selects a physically existing frame while reducing sensitivity to extreme candidate values.

\subsubsection{Multi-Criteria Selection (S3)}
The multi-criteria strategy evaluates each validated candidate using release speed, release angle, and release height. Let $k \in \{s,\alpha,h\}$ index these release parameters, with
$p_{s,r}=s_r$, $p_{\alpha,r}=\alpha_r$, and $p_{h,r}=h_r$.
The multi-criteria score \(M(r)\)  for candidate frame $r$ is defined as:
\begin{equation}
M(r)=
\sum_{k\in\{s,\alpha,h\}}
\beta_k
\exp\left(
-\frac{(p_{k,r}-\tilde{p}_k)^2}{2\sigma_k^2}
\right),
\end{equation}
where $\tilde{p}_k$ is the median of parameter $k$ across the validated candidates, \(\sigma_k\) is the corresponding scale estimate, and \(\beta_k\) controls its relative importance. In this work, the weights for speed, angle, and height are set to \(0.50\), \(0.30\), and \(0.20\), respectively. The final release frame is then selected as the validated candidate with the highest multi-criteria score:
\[
r^*=\arg\max_{r\in\mathcal{C}} M(r).
\]

\section{Experiments and Results Analysis}

\subsection{Dataset and Experimental Setup}
The experimental evaluation was conducted with the same dataset used in~\cite{hasen2026hammer}, which was collected by the Finnish Institute of High Performance Sport (KIHU) between 2017 and 2024. The recordings were captured using a dual-camera setup (side and back views) at frame rates of 240 fps and 180 fps with a resolution of $1920 \times 1080$. The reconstructed 3D trajectories used in this work are obtained using the pipeline introduced in~\cite{hasen2026hammer}.

The proposed release frame detection method is evaluated through the performance and biomechanical consistency of release parameters and distance estimation accuracy. Specifically, each detected release frame is used to estimate the release parameters, including: release speed (m/s), release angle (degrees), and release height (m). These parameters are then used to compute the predicted throw distance using a physics-based trajectory model used in~\cite{hasen2026hammer}. The predicted distance \(d_n^{\text{pred}}\) is compared against the ground-truth measured distance \(d_n^{\text{GT}}\) to compute the error.  The evaluation metrics we used are mean absolute error (MAE), median absolute error (MedAE), and mean absolute percentage error (MAPE): 
\begin{equation}
\text{MAE}=\frac{1}{N}\sum_{n=1}^{N}
\left|d_n^{\text{pred}}-d_n^{\text{GT}}\right|,
\end{equation}
\begin{equation}
\text{MedAE}=
\operatorname{median}
\left(
\left|d_n^{\text{pred}}-d_n^{\text{GT}}\right|
\right),
\end{equation}
\begin{equation}
\text{MAPE}=
\frac{100}{N}\sum_{n=1}^{N}
\left|
\frac{d_n^{\text{pred}}-d_n^{\text{GT}}}{d_n^{\text{GT}}}
\right|,
\end{equation}
where \(n\) indexes the evaluated throws, and \(N\) is the total number of throws.

In addition, we perform a comprehensive ablation analysis to assess the contribution of each signal using the following signal combinations: C1 = $\phi_1$, C2 = C1 + $\phi_2$, C3 = C2 + $\phi_3$, and C4 = C3 + $\phi_4$. Three candidate selection strategies S1-S3 were evaluated for each configuration. This enables a deeper understanding of which cues are most informative for practical automatic release frame detection and how they influence the overall system performance.

\subsection{Overall Performance Analysis}

\begin{table}[t]
\centering
\caption{Performance comparison across signal configurations and candidate selection strategies.}
\label{tab:overall_results}
\begin{tabular}{llccc}
\hline
\textbf{Conf.} & \textbf{Strategy} & \textbf{MAE (m)} & \textbf{MedAE (m)} & \textbf{MAPE (\%)} \\
\hline
C1 & S1 & 5.16 & 4.47 & 7.31 \\
C1 & S2 & 5.19 & 4.45 & 7.34 \\
C1 & S3 & 5.13 & 4.44 & 7.26 \\
\hline
C2 & S1 & 5.11 & 4.41 & 7.24 \\
C2 & S2 & 5.13 & 4.44 & 7.27 \\
C2 & S3 & 4.72 & 4.34 & 6.69 \\
\hline
C3 & S1 & 4.49 & 4.00 & 6.37 \\
C3 & S2 & 4.39 & 4.42 & 6.21 \\
C3 & S3 & 4.73 & 3.75 & 6.70 \\
\hline
C4 & S1 & \textbf{4.37} & 4.06 & \textbf{6.19} \\
C4 & S2 & 4.38 & 4.42 & 6.20 \\
C4 & S3 & 4.71 & \textbf{3.75} & 6.67 \\
\hline
\end{tabular}
\end{table}

The quantitative results for different variants of the proposed MS-RFD framework are summarized in Table~\ref{tab:overall_results}. 
The best result is obtained by the full MS-RFD configuration using weighted aggregation (C4-S1), which achieves an MAE of \(4.37\)~m, a MedAE of \(4.06\)~m, and a MAPE of \(6.19\%\). The representative-frame based selection strategy under the same full configuration (C4-S2) gives nearly the same performance, with an MAE of \(4.38\)~m and a MAPE of \(6.20\%\). The ablation study shows a clear improvement as additional release cues are introduced. The speed-only configuration C1 provides a strong initial baseline, and adding angular transition information in C2 improves the candidate ranking by incorporating directional-change evidence. A more substantial improvement is observed, when radial expansion is introduced in C3, indicating that outward motion is a particularly important cue for separating true release from continued rotation. The final C4 configuration adds post-release linearity as an additional consistency cue. Although the gain from C3 to C4 is smaller, this cue improves the physical interpretation of the selected release frame by favoring candidates followed by a locally linear trajectory segment. 

\begin{figure}[tb]
  \centering
  \includegraphics[width=8.5cm]{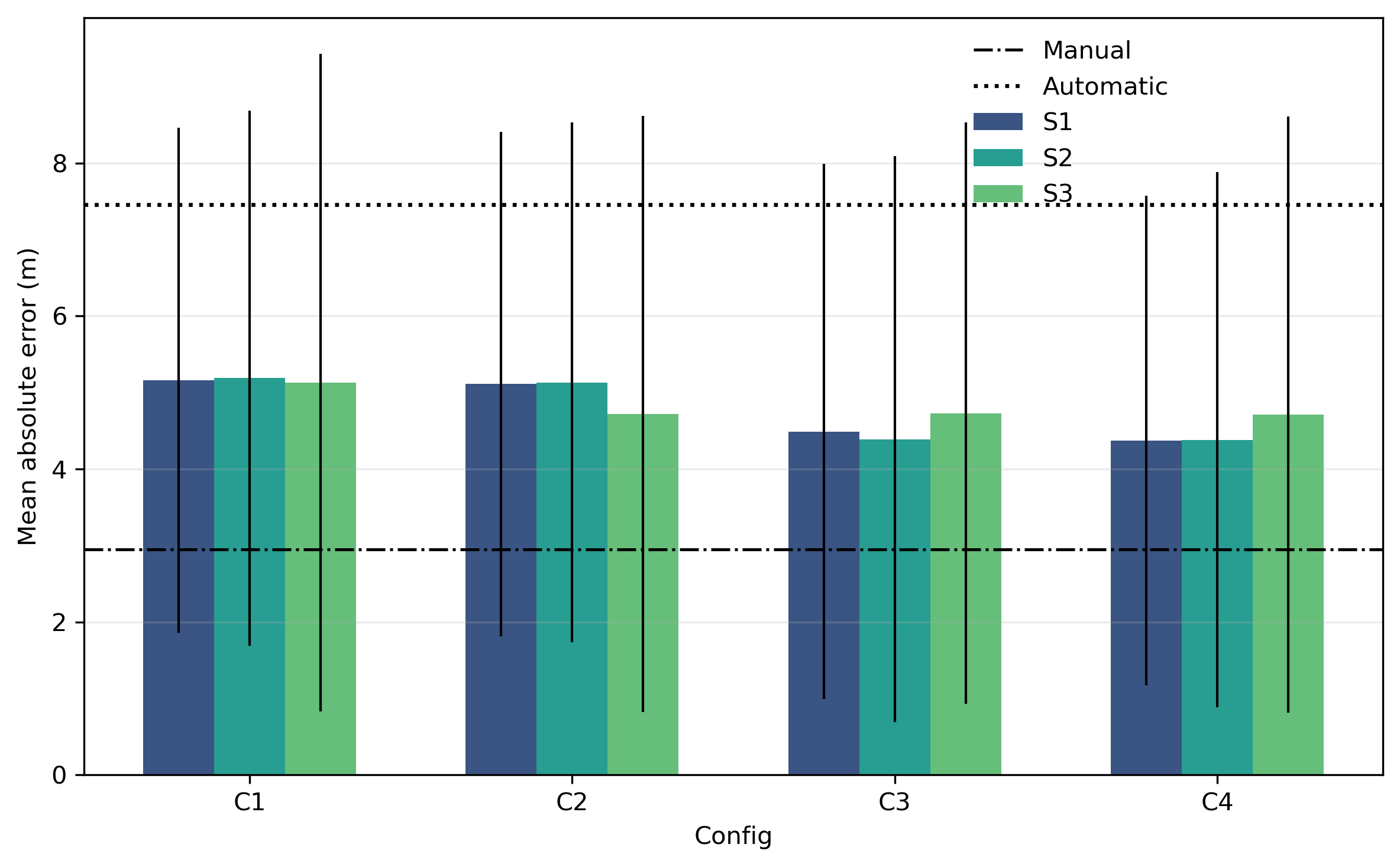}
  \caption{Mean absolute error (MAE) comparison across the four signal configurations and three candidate selection strategies. The bars represent the average error for each strategy, while the vertical error bars indicate the per-throw error variability. The dashed and dotted horizontal lines show the MAE of the previous manual and automatic baselines, respectively.}
  \label{fig:MAE1}
\end{figure}

Fig.~\ref{fig:MAE1} illustrates the MAE comparison across the four signal configurations and shows the overall reduction in error from the speed-only setting toward the full C4 configuration. The figure also shows a comparison against the previous manual and automatic baselines from \cite{hasen2026hammer}. Compared with the previous automatic baseline, which obtains an MAE of \(7.46\) m, C4-S1 reduces the error by approximately \(41.4\%\). The manual baseline still provides the lowest error, as expected, but the proposed method closes a substantial part of the gap between manual release frame selection and automatic detection.

\begin{table*}[t]
\centering
\caption{Per-throw comparison of predicted throwing distances and absolute errors for the Manual baseline, previous Automatic baseline, and the proposed C4 selection strategies.}
\label{tab:c4_results}
\resizebox{0.80\textwidth}{!}{%
\footnotesize
\begin{tabular}{cc|cccc|cccccc}
\hline
\multirow{2}{*}{\textbf{ID}} & \multirow{2}{*}{\textbf{GT}} & \multicolumn{2}{c}{\textbf{Manual Ref.}} & \multicolumn{2}{c}{\textbf{Automatic Ref.}} & \multicolumn{2}{c}{\textbf{C4-S1}} & \multicolumn{2}{c}{\textbf{C4-S2}} & \multicolumn{2}{c}{\textbf{C4-S3}} \\
\cline{3-12}
& & \textbf{Pred.} & \textbf{Error} & \textbf{Pred.} & \textbf{Error} & \textbf{Pred.} & \textbf{Error} & \textbf{Pred.} & \textbf{Error} & \textbf{Pred.} & \textbf{Error} \\
\hline
11 & 70.26 & 73.22 & 2.96 & 72.38 & 2.12 & 77.10 & 6.84 & 76.70 & 6.44 & 76.70 & 6.44 \\
12 & 73.51 & 74.49 & 0.98 & 76.79 & 3.28 & 72.85 & 0.66 & 74.52 & 1.01 & 71.39 & 2.12 \\
13 & 68.25 & 69.87 & 1.62 & 82.22 & 13.97 & 69.14 & 0.89 & 69.89 & 1.64 & 68.93 & 0.68 \\
21 & 70.19 & 79.66 & 9.47 & 87.89 & 17.70 & 79.27 & 9.08 & 79.75 & 9.56 & 74.53 & 4.34 \\
22 & 70.68 & 69.51 & 1.17 & 62.08 & 8.60 & 71.43 & 0.75 & 75.10 & 4.42 & 67.22 & 3.46 \\
23 & 71.03 & 71.34 & 0.31 & 73.21 & 2.18 & 66.97 & 4.06 & 74.27 & 3.24 & 62.21 & 8.82 \\
24 & 69.89 & 66.18 & 3.71 & 74.64 & 4.75 & 77.19 & 7.30 & 78.08 & 8.19 & 75.97 & 6.08 \\
25 & 70.36 & 60.12 & 10.24 & 45.62 & 24.74 & 61.57 & 8.79 & 64.17 & 6.19 & 56.05 & 14.31 \\
26 & 71.07 & 73.30 & 2.23 & 65.83 & 5.24 & 75.67 & 4.60 & 76.25 & 5.18 & 76.01 & 4.94 \\
27 & 72.76 & 73.09 & 0.33 & 79.63 & 6.87 & 78.94 & 6.18 & 79.64 & 6.88 & 78.33 & 5.57 \\
31 & 70.07 & 75.09 & 5.02 & 77.72 & 7.65 & 81.17 & 11.10 & 80.78 & 10.71 & 79.70 & 9.63 \\
32 & 70.17 & 70.37 & 0.20 & 58.48 & 11.69 & 73.50 & 3.33 & 73.67 & 3.50 & 73.74 & 3.57 \\
33 & 72.56 & 68.49 & 4.07 & 81.06 & 8.50 & 73.61 & 1.05 & 72.19 & 0.37 & 72.19 & 0.37 \\
34 & 71.32 & 70.05 & 1.27 & 74.29 & 2.97 & 68.63 & 2.69 & 70.06 & 1.26 & 70.06 & 1.26 \\
35 & 73.82 & 73.43 & 0.39 & 69.48 & 4.34 & 78.16 & 4.34 & 78.80 & 4.98 & 77.57 & 3.75 \\
36 & 65.28 & 68.70 & 3.42 & 71.16 & 5.88 & 64.83 & 0.45 & 66.12 & 0.84 & 63.66 & 1.62 \\
37 & 69.36 & 66.68 & 2.68 & 83.22 & 13.86 & 67.09 & 2.27 & 69.26 & 0.10 & 66.19 & 3.17 \\
\hline
\textbf{MAE} & -- & -- & \textbf{2.95} & -- & \textbf{7.46} & -- & \textbf{4.37} & -- & \textbf{4.38} & -- & \textbf{4.71} \\
\hline
\end{tabular}%
}
\end{table*}

\begin{figure}[tb]
  \centering
  \includegraphics[width=8.5cm]{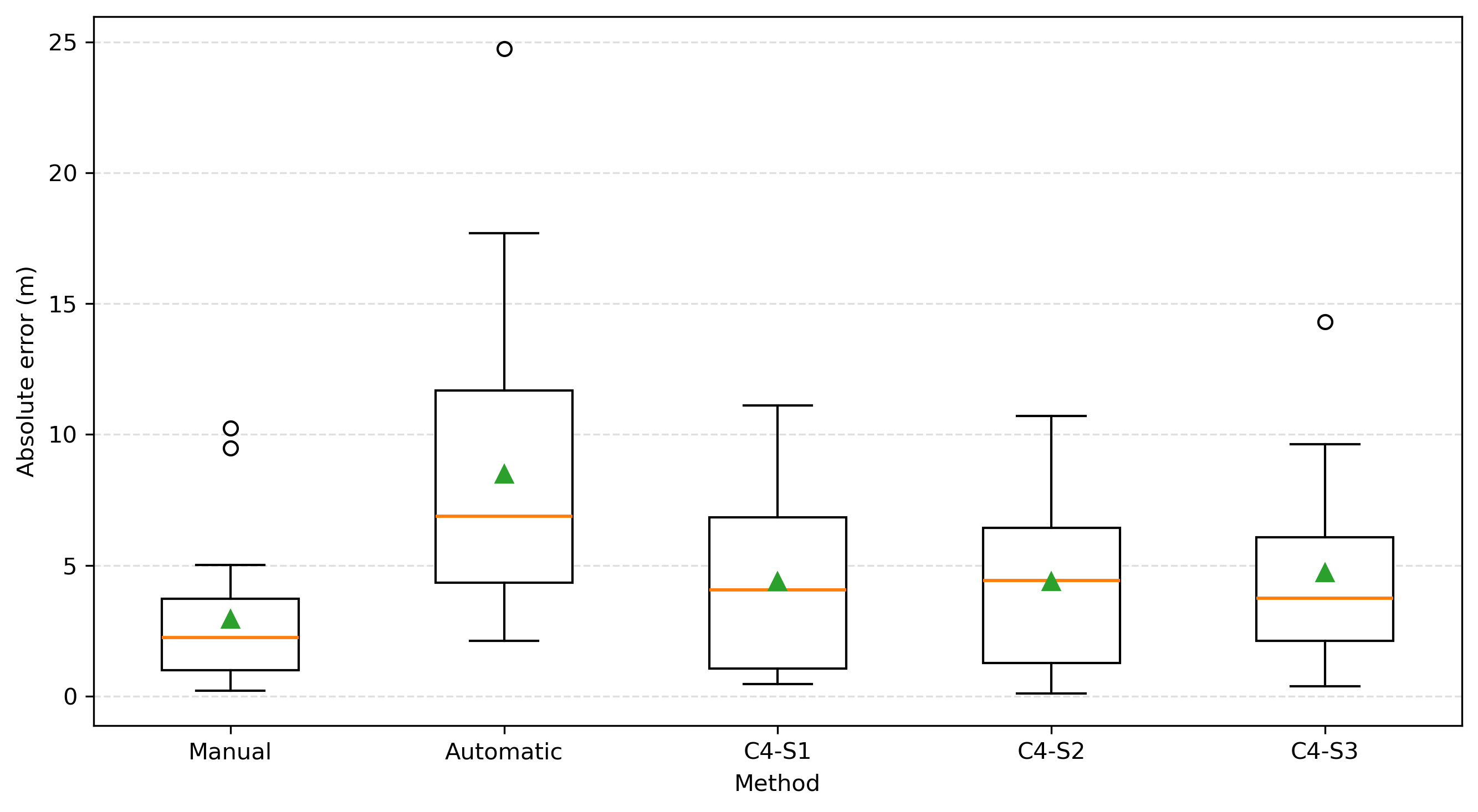}
  \caption{Distribution of absolute distance errors for the manual baseline, previous automatic baseline, and the three C4 selection techniques. The boxplots show the median, interquartile range, mean, and outliers.}
  \label{fig:AED1}
\end{figure}


\subsection{Per-Throw Error Analysis}

Table~\ref{tab:c4_results} presents the detailed per-throw predictions for the full C4 configuration, together with the manual and automatic baselines. The manual baseline achieves the lowest MAE of \(2.95\) m, which is expected because it benefits from human interpretation of the release event. Among automatic methods, the proposed C4 variants substantially outperform the previous automatic baseline. The automatic baseline obtains an MAE of \(7.46\) m, while C4-S1 and C4-S2 reduce the automatic error to approximately \(4.38\) m. Fig.~\ref{fig:AED1} shows the distribution of absolute errors for the manual baseline, previous automatic baseline, and C4, highlighting that C4 reduces the error spread compared with the previous automatic method.

Several throws are estimated with high accuracy. For example, IDs 12, 13, 22, 33, 36, and 37 achieve low errors. These cases indicate that MS-RFD is effective when the reconstructed trajectory contains a clear rotational-to-ballistic transition. The largest remaining errors occur for IDs 31, 25, 24, 21, and 11. These throws dominate the MAE and suggest that the main limitation is not the average behavior of the detector, but a small number of difficult trajectories where the release transition is either noisy, temporally ambiguous, or affected by reconstruction uncertainty. 



The comparison against the manual baseline is also informative. The proposed method does not yet match manual selection, but it closes a large part of the gap between previous automatic release detection and manual method. 

\subsection{Qualitative Analysis}

\begin{figure}[tb]
  \centering
  \includegraphics[width=8.5cm]{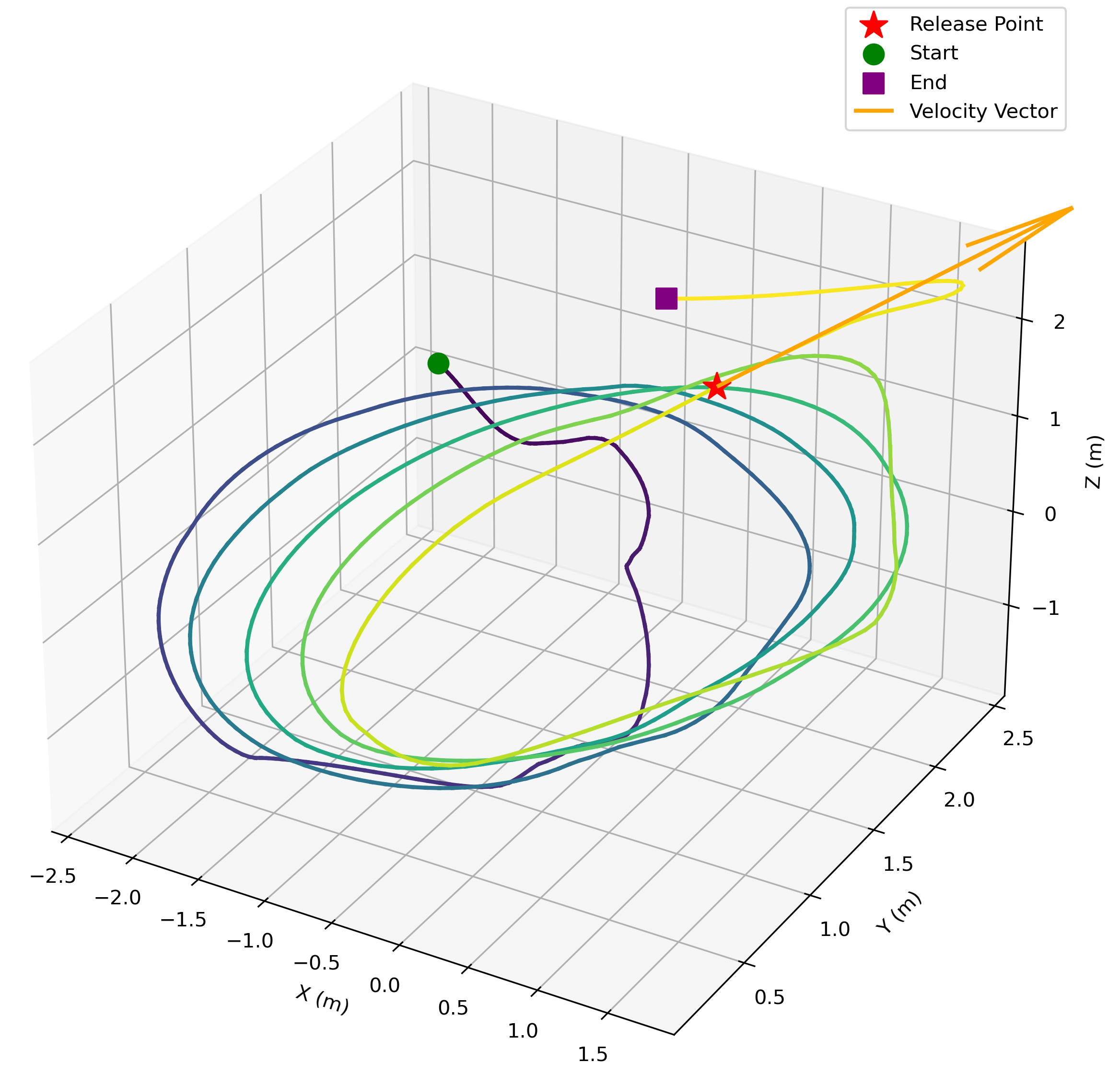}
  \caption{Reconstructed 3D hammer trajectory showing the estimated release point corresponding to the detected release frame.}
  \label{fig:trajectory_example}
\end{figure}

Fig.~\ref{fig:trajectory_example} shows an example of a reconstructed 3D hammer trajectory and the estimated release point corresponding to the detected release frame, highlighting the transition from rotational motion to linearity using our MS-RFD method, where speed, angular change, radial expansion, and future trajectory linearity are combined to identify a physically plausible release frame.

\section{Conclusion}

This paper presented MS-RFD, a multi-signal framework for automatic release-frame detection in hammer throw using reconstructed trajectories. The method combines speed dynamics, angular transition, radial expansion, and post-release linearity to identify physically plausible release frames. The results show that explicitly modeling hammer-throw mechanics improves automatic release detection. The ablation results further show that each signal contributes complementary information, with the strongest performance obtained when all cues are combined. The full C4 configuration achieved the best automatic performance, reducing MAE from \(7.46\)m for the previous automatic baseline to \(4.37\) m. Although the proposed method does not yet match manual release-frame selection, it substantially closes the gap between previous automatic detection and human-guided analysis. The current study has some limitations. First, MS-RFD depends on the quality of the reconstructed 3D trajectory, so errors from detection, calibration, triangulation, or smoothing can affect the trajectory near release and influence the estimated release parameters and predicted distance. In addition, the method uses fixed fusion weights and temporal windows, which were kept constant across experiments rather than optimized for individual athletes or recording setups. Furthermore, the evaluation is mainly based on downstream throwing-distance error rather than direct frame-level release annotations.

\section*{Acknowledgment}

This work was part of Finland's Ministry of Education and Culture’s Doctoral Education Pilot under Decision No. VN/3137/2024-OKM-6 (The Finnish Doctoral Program Network in Artificial Intelligence, AI-DOC).

\bibliographystyle{IEEEtran}
\bibliography{ref}

\end{document}